\def\year{2020}\relax
\documentclass[letterpaper]{article} 
\usepackage{aaai20}  
\usepackage{times}  
\usepackage{helvet} 
\usepackage{courier}  
\usepackage[hyphens]{url}  
\usepackage{graphicx} 
\usepackage{amsmath}
\usepackage[caption=false,font=footnotesize]{subfig}
\usepackage{ gensymb }
\usepackage{graphicx}  
\title{Robot Motion Risk Reasoning Framework}

\author{Xuesu Xiao, Jan Dufek, Robin R. Murphy\\ 
Department of Computer Science and Engineering\\Texas A\&M University\\ 
College Station, Texas 77843\\
xiaoxuesu@tamu.edu, dufek@tamu.edu, robin.r.murphy@tamu.edu 
}

\begin{document}

\maketitle

\begin{abstract}
This paper presents a formal and comprehensive reasoning framework for robot motion risk, with a focus on locomotion in challenging unstructured or confined environments. Risk which locomoting robots face in physical spaces was not formally defined in the robotics literature. Safety or risk concerns were addressed in an ad hoc fashion, depending only on the specific application of interest. Without a formal definition, certain properties of risk were simply assumed but ill-supported, such as additivity or being Markovian. The only contributing adverse effect being considered is related with obstacles. This work proposes a formal definition of robot motion risk using propositional logic and probability theory. It presents a universe of risk elements within three major risk categories and unifies them into one single metric. True properties of risk are revealed with formal reasoning, such as non-additivity or history-dependency. Risk representation which encompasses risk effects from both temporal and spatial domain is presented. The resulted risk framework provides a formal approach to reason about robot motion risk. Safety of robot locomotion could be explicitly reasoned, quantified, and compared. It could be used for risk-aware planning and reasoning by both human and robotic agents. 
\end{abstract}

\section{Introduction}
Robots are always used as a replacement for humans in dirty, dull, or dangerous places \cite{murphy2000introduction}. Those environments usually pose dangers or are inaccessible to human agents and are therefore considered to be too ``risky'' for human venture. However, deploying robots into those spaces also inevitably causes risk for the robots. Either being tele-operated or autonomous, the robot's own safety may be endangered by their interaction with the challenging environments. The usages of unmanned vehicles in situations such as Urban Search And Rescue (USAR), nuclear operations, disaster robotics \cite{murphy2014disaster}, etc., are examples where the unstructured or confined environments pose extreme risk for the robots tasked to operating in them. Although consequences caused by accidents may not be fatal, the cost to mend them is also expensive, in terms of mission-criticalness, delay in response, economic loss, damage to the environment, etc. 

\begin{figure}[]
\centering
\includegraphics[width=0.5\columnwidth]{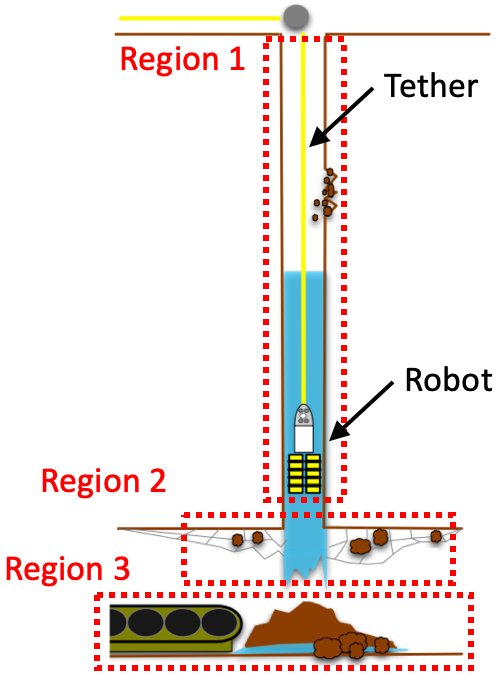}
\caption{Mine Disaster Borehole Entry \cite{murphy2009mobile}}
\label{fig::BE_3regions}
\end{figure}

One example of robots being deployed to unstructured or confined environments and face risk is victim search in Crandall Canyon Mine (Utah) response in 2007. Borehole entry (Fig. \ref{fig::BE_3regions}) utilizes the small boreholes which are drilled into the mine into what is expected to be the affected area and the robot uses those as entry point. The idea in the borehole scenario is to insert a small robot into the borehole, drop the robot to the floor, and explore the affected area \cite{murphy2009mobile}. In region 1 (the borehole area), multiple risk sources exist at the same time: due the small clearance of the borehole, it is very likely that the robot will get jammed. Due to the lack of casing of the borehole, falling rocks may damage the robot. Drilling foam, water, and debris may cause system malfunction as well. The vertically hanging robot might spin and therefore lose controllability and mobility. In region 2, the transition from the borehole to the mine, mesh roof exists as the existent structure of the mine. The robot faces risk due to the mesh roof interfering with hole exit and reentry. Because the robot is tethered, risk of robot tether getting tangled with the mesh roof is significant. Furthermore, the transition from vertical mobility to operating on mine floor also requires extra effort and induces risk. Region 3 is the inside of the mine, where extra risk sources appear after the disaster. Terrain may be unstable due to running water and mud, causing the robot getting trapped and stuck. The robot also has to traverse soft drill tailings and foam, or even equipment, before reaching the mine floor. Any of those can pose risk to the robot. Lastly, while locomoting in region 3, robot tether is still being extended or retracted, interacting with the borehole (region 1) and the mesh roof in the transition into the mine (region 2). Risk of tether entanglement still exists. 

Due to the variety of existing risk sources in borehole entry, the robot failed at Crandall Canyon Mine (Utah) in all four runs during the response in 2007. The reasons for the failure are: (1) the lowering system failed, (2) the robot encountered a blockage in the borehole, (3) The robot had to be removed to clean the lens from the buildup of water, debris, and drilling foam, (4) the robot tether was entangled with the mesh and the robot was trapped on the way back. After it was freed, the robot was lost when the tether finally broke due to the actively eroding borehole's severe washout and large boulders. 

How the variety of risk sources in the unstructured or confined environment contribute to the high failure rate (100\%) of the deployment of a sophisticated robot system designed and engineered for those purposes motivates a fundamental reasoning framework which includes a formal definition of risk locomoting robot faces in unstructured or confined environments and an explicit representation to reason, quantify, and compare risk.  

Motivated by this, the paper formally defines robot motion risk using propositional logic and probability theory. The use of those formal methods reveals the dependencies of risk a robot faces at a certain point on a path on the history leading to this point. It also articulates how risks at individual steps are combined into risk of executing the entire path. These are in contrast to the conventionally assumed but ill-founded additive or Markovian properties. The proposed framework also captures the combined effect from different risk sources during robot locomotion, other than one ad hoc risk (mostly obstacle-related). Building upon a comprehensive universe of risk elements (not only obstacles), a variety of adverse effects which exist in unstructured or confined environments are categorized into locale-dependent, action-dependent, and traverse-dependent risk elements. This approach is formal and therefore general, comprehensive, and objective. The resulted risk framework provides a formal approach to reason and quantify motion risk and serves as a tool for explicit and intuitive comparison between different motion plans. It builds a formal common ground for risk-aware behaviors. This fundamental framework could be used for reasoning by both human and robotic agents. 

\section{Related Work}
\label{sec::related_work}
To the author's best knowledge, a formal definition of risk does not exist in the literature, except being referred to as some negative impact or factor in ad hoc situations. Or it is simply treated as a numerical measure of the severity/negativity related with certain aspects of motion. Due to the lack of a formal definition of what risk is in the robotics literature, the ad hoc risk representation used in existing works has very low fidelity in modeling actual risk and lacks comprehensiveness and accuracy. The two major categories to model risk are (1) explicit ad hoc risk functions and (2) (chance) constraints. 

\subsection{Ad Hoc Risk Functions}
\citeauthor{soltani2004fuzzy} represented the workspace by two risk layers: hazard data and visibility layer. The risk of each state along the path was computed based on the distance to hazard and visibility value using fuzzy logic. The objective function was a weighted sum of the two layers accumulated along the entire path. To the author's best knowledge, this is the only work in the literature that considered more than one risk sources: risk from being close to hazard and risk from having low visibility. A similar approach was taken by \citeauthor{de2011minimum}, where a risk map was generated based on ground orography treated as obstacle. \citeauthor{vian1989trajectory} presented the idea of risk index for any particular location (state) and assumed risk to be a function of location only. \citeauthor{zabarankin2002optimal} based their risk representation on the same risk index idea, whose value was proportional to risk factor and reciprocal to squared distance to threat. Risk caused by multiple threats were summed and this accumulated value was integrated along the path. \citeauthor{gu2006comprehensive} further proposed an accumulative parametrized function based on distances to multiple threats. The set of functional parameters were set manually. \citeauthor{feyzabadi2014risk} used a similar distance-based function to represent state-dependent risk in its experiment. Data-driven approaches to predict potential risk of a certain state could also be seen in prior works. In the field of Autonomous Underwater Vehicles (AUVs), \citeauthor{pereira2011toward} defined risk as a function of state location with ship occurrences averaged over time domain since historical Automated Identification System (AIS) data was available. In traffic planing, \citeauthor{krumm2017risk} utilized historical traffic data to predict crash probability to represent the risk associated with driving through each corresponding highway segment. 



From the above-mentioned explicit ad hoc risk representation, distance to closet threat is the main risk element considered, in the form of hazard, orography, radar, obstacle, etc. Although it is naturally assumed that being closer to threat brings more risk, a formal definition of what risk is is still missing. The actual meaning of the risk function is unclear. \cite{soltani2004fuzzy} is the only work that considered multiple risk elements and the approach to combine them was weighted sum based on human heuristics. All those previous works explicitly represented risk as a function of current state alone. They also assumed that risk is additive: risk associated with an entire path is a simple addition of all the risks at each individual states. However, the justification of the state-dependency and additivity remains missing. 

\subsection{(Chance) Contraints}
Other than modeling risk using an explicit ad hoc risk function, other works addressed risk as constraint violation or probability of constraint violation (chance constraints). 

\subsubsection{MDP: Reward with (Chance) Constraints}
A popular approach to handle reward and risk is to use (PO)MDP. As standard MDP inherently contains reward but not risk, researchers have looked into representing risk as negative reward (penalty) or constraints (C-POMDP) with unit cost for constraint violation. \citeauthor{pereira2013risk} modeled constraints as penalties on the reward by subtracting penalty from reward function. \citeauthor{undurti2010online} modeled risk as constraint violation in MDP framework. Here, risk was treated as unit cost incurred when a hard constraint on the system would be violated. Going beyond negative reward and unit cost for constraint violation, Chance Constrained Partially Observable Markov Decision Process (CC-POMDP) was proposed by \citeauthor{santana2016rao}, which was based on a bound on the probability (chance) of some event happening during policy execution. 

\subsubsection{(Chance) Constrained RMPC}
Robust Model Predictive Control (RMPC) is an alternative to MDP-based methods. \citeauthor{luders2010chance} proposed a chance-constrained rapidly-exploring random tree (CC-RRT) approach, which used chance constraints to guarantee probabilistic feasibility at each time step. \citeauthor{luders2013robust} expanded this approach to guarantee probabilistic feasibility over entire trajectories (CC-RRT*). Other works emphasized on risk allocation, i.e., to allocate more risk for more rewarding actions. \citeauthor{ono2008iterative}  used a two stage optimization scheme with the upper stage optimizing risk allocation and lower stage calculating optimal control sequence that maximizes reward, named Iterative Risk Allocation (IRA). \citeauthor{vitus2011feedback} also used risk allocation and feedback controller optimization to reduce conservatism and improve performance. Risk was also represented as a bounded probability of constraint violation. 

All the approaches in the literature, which modeled risk as chance constraints, only focused on risk, or constraints, caused by collision with obstacles. With the system modeled within Cartesian space, the constraints of the dynamic system were formulated as no intersection between the robot trajectory and obstacles in the environment at each time step. Being modeled only in a geometric point of view, approaches to model risk caused by any other sources than obstacles were overlooked, e.g., robot motor overheat, getting stuck in granular environments, etc. Using chance constraints, risk was only a manually defined bound or threshold of constraint violation probability. Furthermore, the temporal or spatial (multiple obstacles) dependencies of constraint violation probability were either assumed to be independent or relaxed using ellipsoidal relaxation technique or Boole's inequality. For example, the probability of constraint violation at this time step was assumed or relaxed to be only a function of $x_t$, the state at this time. These two methods, especially when residing only in Cartesian space, neglected the important dependencies on the motion history and the rough approximation introduced significant conservatism. 

To summarize existing works, explicit risk functions were ad hoc and assumed ill-founded properties of risk, such as additive and Markovian. (Chance) constrained methods only addressed risk within a probability bound, and either assumed temporal and spatial independence or used conservative relaxation methods, which is equal to assuming independence. Furthermore, both methods only focused on obstacle-related risk. This work proposes a formal definition of robot motion risk using propositional logic and probability theory. The proposed framework unifies most existing robot motion risk sources into one single metric to explicitly represent risk levels of different motion plans. The formal methods reveal contradicting risk properties: non-additivity and history-dependency. This definition and representation provide an intuitive approach to compare risk of different paths to improve robot locomotion safety. As a new tool to reason and quantify safety, this formal risk reasoning framework could be used to compare grounded risk-awareness. 

\section{Formal Risk Definition}
This work considers motion risk for mobile robots executing a preplanned path. Risk in terms of a sequence of motion (path) is formally defined as \emph{the probability of the robot not being able to finish the path}. 

Before reasoning about risk of executing a path, the workspace of the robot is firstly defined based on tessellation of the Cartesian space, either in 2D or 3D, depending on where the robot resides. Each tessellation is either a viable (e.g. free) or unviable (e.g. occupied) state for the robot to locomote. A feasible path plan $P$ is defined to be an ordered sequence of viable tessellations, called \emph{states} and denoted as $s_i$:

\begin{center}
$P = \{s_0, s_1, ..., s_n\},~|| s_i - s_{i-1}||_2 \leq r_c, \forall 1 \leq i \leq n$, 
\end{center}

where $r_c$ is the maximum distance between two consecutive states for the path to be feasible. 

A state on the path is \emph{finished} by the robot reaching the state within an acceptable tolerance and ready to move on to the next state. A state is \emph{not finished} due to two main reasons: the robot crashes or gets stuck. In order to finish the path of $n$ states, the robot faces $r$ different risk elements. which will possibly cause not finishing the path (crash or getting stuck). Three types of events are defined: 

\begin{itemize}
\item $F$ -- the event where the robot finishes path $P$
\item $F_i$ -- the event where the robot finishes state $i$
\item $F_i^k$ -- the event where risk $k$ does not cause a failure at state $i$
\end{itemize}

The reasoning about motion risk is based on three assumptions, which are expressed by propositional logic: 
\begin{enumerate}
	\item Path is finished only when all states are finished: 
	\begin{center}
	$F = F_n \cap F_{n-1} \cap ... \cap F_1 \cap F_0$
	\end{center}
	
	\item A state is finished only when all risk elements do not cause failure: 
	\begin{center}
	$F_i = F_i^1 \cap F_i^2 \cap ... \cap F_i^{r-1} \cap F_i^r$
	\end{center}
	
	\item Finish or fail a state because of one risk element is conditionally independent of finish or fail that state because of any other risk element, given the history leading to the state:  
	\begin{center}
	$(F_i^1 \vert \bigcap \limits_{j=0}^{i-1} F_j) \perp \!\!\! \perp (F_i^2 \vert \bigcap \limits_{j=0}^{i-1} F_j) \perp \!\!\! \perp ... \perp \!\!\! \perp (F_i^{r-1} \vert \bigcap \limits_{j=0}^{i-1} F_j) \perp \!\!\! \perp (F_i^r \vert \bigcap \limits_{j=0}^{i-1} F_j)$
	\end{center}
\end{enumerate}

As complement of the formal risk definition proposed by this work, the probability of the robot \emph{being} able to finish the path could be written as $P(F)$: 

\begin{equation}
\begin{split}
P(F) 
&= P(F_n\cap F_{n-1} \cap ... \cap F_0) \\
&= P(F_n \vert F_{n-1} \cap ... \cap F_0) \cdot ... \cdot P(F_1 \vert F_0) \cdot P(F_0)\\
&= \prod_{i=0}^{n} P(F_i \vert \bigcap_{j=0}^{i-1} F_j) \\
& = \prod_{i=0}^{n} P(F_i^1 \cap F_i^2 \cap ... \cap F_i^r \vert \bigcap_{j=0}^{i-1} F_j) \\
& = \prod_{i=0}^{n} P(F_i^1 \vert \bigcap_{j=0}^{i-1} F_j) \cdot ... \cdot P(F_i^r \vert \bigcap_{j=0}^{i-1} F_j) \\
& = \prod_{i=0}^{n} \prod_{k=1}^{r} P(F_i^k\vert \bigcap_{j=0}^{i-1} F_j)
\end{split}
\end{equation}

The first, second, fourth, and fifth equal signs are due to assumption 1, probability chain rule, assumption 2, and assumption 3, respectively. The third and sixth are simply reformulation. Therefore, the formal risk definition, the probability of \emph{not} being able to finish the path, is the probabilistic complement: 

\begin{equation}
\begin{split}
P(\bar{F}) 
&= 1 - P(F) \\
&= 1 - \prod_{i=0}^{n} \prod_{k=1}^{r} P(F_i^k \vert \bigcap_{j=0}^{i-1} F_j) \\
&= 1 - \prod_{i=0}^{n} \prod_{k=1}^{r} (1-P(\bar{F_i^k} \vert \bigcap_{j=0}^{i-1} F_j))
\end{split}
\label{eqn::pfbar}
\end{equation}

In terms of risk representation, the risk of path $P$ is denoted as $risk(P)$ and is equal to $P(\bar{F})$. $P(\bar{F_i^k} \vert \bigcap\limits_{j=0}^{i-1} F_j)$ means the probability of risk $k$ causes a failure at state $i$, given the history of finishing $s_0$ to $s_{i-1}$. It is therefore denoted as the $k$th risk robot faces at state $i$ given that $s_0$ to $s_{i-1}$ were finished: $r_k(\{s_0, s_1, ..., s_i\})$. 

Writing in risk representation form will yield: 

\begin{equation}
risk(P) = 1 - \prod_{i=0}^{n} \prod_{k=1}^{r} (1-r_k(\{s_0, s_1, ..., s_i\}))
\label{eqn::risk_representation}
\end{equation}

This is the proposed robot motion risk framework to quantify the risk of executing the path. In contrast to the traditional additive state-dependent risk representation, the proposed approach gives a probability value in $[0, 1]$ instead of $[0, \infty]$. It does not require the ill-founded additivity assumption for risk. More importantly, the conditional probability in Eqn. \ref{eqn::pfbar} clearly shows the dependency of risk at certain state on the history, not only the state itself. So the risk the robot is facing at a certain point is not only state-dependent, but also depends on the history leading to this state. For example, if the robot takes a very muddy path to come to a muddy state, the probability of getting stuck at this state is high, due to the mud built up on the wheels or tracks in the history. However, if a clean path was taken, risk at this very same state may be minimum, since clean wheels or tracks can easily maintain sufficient traction. 

Despite the dependencies in the temporal domain, conditional independence among different risk elements at a certain state given the history is still assumed. For instance, if the robot will crash to the closest obstacle is independent of if the robot will tip over due to a sharp turn. This independence assumption matches with the intuition when multiple unrelated risk sources are affecting the robot at the same time. Along the direction of the path, risk the robot faces at each individual state is dependent on history (longitudinal dependence), while at each state, the risks caused by different risk elements are independent (lateral independence). 


\section{Risk Universe and Categories}
\begin{figure}[]
\centering
	\includegraphics[width = 1 \columnwidth]{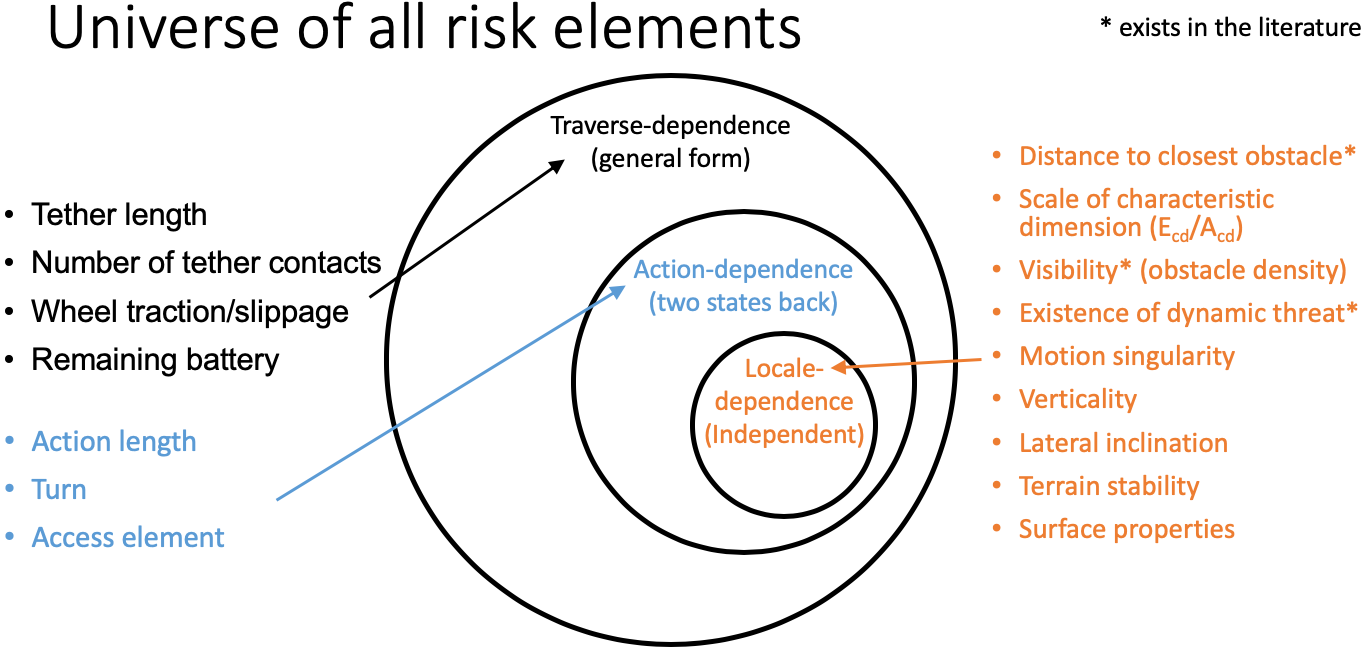}
	\caption{Universe of All Risk Elements}
	\label{fig::universe}
\end{figure}

The formal definition and explicit representation reveal the longitudinal dependence of risk at a certain state on the history. Mathematically speaking, the dependency is on the entire history in general. However, in practice, the dependency of different risk elements may have different depth into the history, e.g. crash to a very close obstacle is only dependent on the closeness of this state to obstacle or crash due to an aggressive turn is only dependent on two states back in the history. In this work, risk elements are divided into three categories: \emph{locale}-dependent, \emph{action}-dependent, and \emph{traverse}-dependent risk elements. Fig. \ref{fig::universe} shows a universe of risk elements, and the categories they belong to. The list is not exclusive. More importantly, the subset/superset relationship between the three categories are displayed: locale-dependence $\subset$ action-dependence $\subset$ traverse-dependence. This section will explain each categories. 

\subsection{Locale-dependent Risk Elements}

Locale-dependent risk is the most special case in history dependence, since its dependency on history could be entirely relaxed. That is: 

\begin{equation}
P(\bar{F_i^k} \vert \bigcap_{j=0}^{i-1} F_j) = P(\bar{F_i^k})
\end{equation}

The word \emph{locale} connotes the meaning of ``location'', ``position'', or where the robot is currently at. It has similar connotation as the concept of ``state'' in (Cartesian) configuration space, but also emphasizes the relationship with the current proximity of the environment. This category of risk elements has been covered in existing literature under the name of ``location'' or ``state'' and was assumed to be the only type of risk elements. This type of traditional risk elements could be evaluated on the state alone, not depending on history.

\subsection{Action-dependent Risk Elements}
Action-dependent risk is a special case of risk's history dependency, between the general traverse-dependence and the most special locale-dependence. The depth of action-dependent risk elements' history dependency is two states back, such that the finishing of the last two states have impact on the risk the robot is facing at the current state: 

\begin{equation}
P(\bar{F_i^k} \vert \bigcap_{j=0}^{i-1} F_j) = P(\bar{F_i^k} \vert F_{i-2} \cap F_{i-1})
\end{equation}

This category of risk elements usually focuses on the transitions between states, including the effort necessary to initiate the transition and the difference between two consecutive transitions. 

\subsection{Traverse-dependent Risk Elements}
Traverse-dependent risk is the general form of risk's history dependency, which encompasses both locale-dependent and action-dependent risk elements. The general form has a full depth of history dependency and looks back to the whole traverse from start leading to the current state. Finishing of all the history states has impact on the finishing of the current state: 
 
\begin{equation}
P(\bar{F_i^k} \vert \bigcap_{j=0}^{i-1} F_j) = P(\bar{F_i^k} \vert F_{i-1} \cap F_{i-2} \cap ... \cap F_1 \cap F_0) 
\end{equation}

\section{Risk Representation Results}
With the formal definition of risk as the probability of the robot not being able to finish the path, along with three categories of, locale-dependent, action-dependent, and traverse dependent, risk elements, this section presents a step-by-step example explaining given a feasible path in an unstructured or confined environment, how risk is represented as a numerical probabilistic value to reason about the risk the robot faces at each individual state and along the entire path. Other examples are also provided and compared with conventional additive state-dependent risk representation. 

Eqn. \ref{eqn::risk_representation} is the basic formulation for risk representation. The risk of executing the entire path $P$ is evaluated based on the contributions each individual risk element (risk element $1$ to $r$) has at each individual state on the path (state $0$ to $n$). Conventional risk representation approaches assumed additivity of risk, i.e. the risk of an entire path is the summation of the risks of individual states. The additivity, however, is not well supported. The result of the conventional risk representation was a risk index in $[0, \infty]$, whose definition and meaning remain unclear. They also only considered locale-dependent risk, ignoring all the dependencies on the finishing of history states, i.e. action-dependent and traverse-dependent risk elements. The proposed approach is grounded on a formal risk definition and uses propositional logic and probability theory to combine the individual effect of risk at state and risk caused by individual risk element into risk of a path. It also considers action-dependent and traverse-dependent risk elements, in addition to locale-dependent risk elements. The output of the proposed risk representation is a risk value exactly as the probability of the robot not being able to finish the path. 

As shown in Eqn. \ref{eqn::risk_representation}, given a state $s_i$, the risk contributed by one risk element $r_k$ is in general dependent on the history states on the traverse $s_0, s_1, ..., s_{i-1}$. The value of this particular risk, as the probability of this risk element $k$ causes failure at this state $i$, could be computed either empirically or theoretically. In the absence of an theoretical approach to compute the probability value, this risk could be calculated based on the extent of the adverse property, e.g. being closer to obstacle, making sharper turn, and having more tether contact points \cite{xiao2018motion} will have a higher probability of failure at this state. Those probability values could be empirically determined. 

\begin{figure}[]
\centering
	\includegraphics[width = 0.5 \columnwidth]{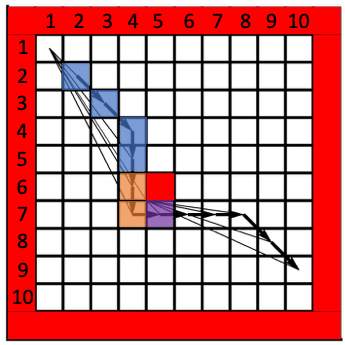}
	\caption{Example Environment and Path for Risk Representation}
	\label{fig::risk_representation_example2}
\end{figure}

In order to illustrate risk representation, this section uses a tethered UAV \cite{xiao2019autonomous} as example, and three representative risk elements are chosen in order to cover all three risk categories and maintain simplicity at the same time. The three example risk elements are distance to closest obstacle as locale-dependent risk element, turn as action-dependent risk element, and number of contact points as traverse-dependent risk element. The workspace is based on the tessellation of 2D Cartesian space, surrounded by obstacles and one extra obstacle in the middle. The workspace and example path to be evaluated is shown in Fig. \ref{fig::risk_representation_example2}. For better illustration, other than the index of each state ($0-11$), the subscript also corresponds to the index of rows and columns of the state in the 2D occupancy grid (first and second column in Tab. \ref{tab::risk_representation_table}). 

The path starts from the upper left corner ($s_{22}$) and ends at the lower right corner ($s_{910}$). For each state, all three risk elements are evaluated. Risk caused by distance to closest obstacle is only based on the current locale alone, where the current state locates. The distance value to the closest obstacle is mapped into a risk value empirically, denoting the probability of not being able to finish this state (third column in Tab. \ref{tab::risk_representation_table}). Risk caused by turn is action-dependent, so two states back need to be investigated. Based on the difference of the two consecutive actions, a risk value is empirically assigned (forth column in Tab. \ref{tab::risk_representation_table}). For traverse-dependent number of tether (shown as thin black lines) contact points, by looking back at the entire traverse, the number of contact points could be determined \cite{xiao2018motion}. Here we simply assign $0.03$ probability of not finish to states with one contact point and $0$ to those that don't have one (fifth column in Tab. \ref{tab::risk_representation_table}). 

\begin{table}[h!]
\centering
\includegraphics[width = 1 \columnwidth]{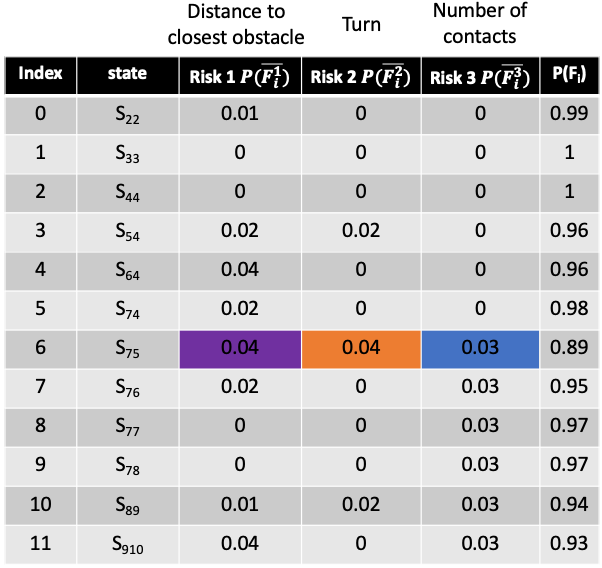}
\caption{Risk Representation for Individual States and Risk Elements (Probability Conditions are Omitted in the Notation for Brevity)}
\label{tab::risk_representation_table}
\end{table}

The sixth state $s_{75}$ on the path is chosen as example to illustrate how the three risk values from three risk elements are computed (color-coded in Fig. \ref{fig::risk_representation_example2} and Tab. \ref{tab::risk_representation_table}). Due to the closeness to the obstacle in the middle, the risk of collision and therefore not being able to finish the state is 0.04. This risk value only needs to be evaluated by the purple block alone, the current state itself. By looking back two states into history ($s_{64}$ and $s_{74}$), the robot needs to move down first and then make a sharp 90\degree~turn to move right. Due to the sharpness of the turn, there is 0.04 probability that the robot cannot make the turn and reach $s_{75}$. Note that $s_{75}$ should also be in orange, but due to the overlap with purple the orange is omitted. In terms of contact point, the entire traverse needs to be taken into account (blue blocks), in order to determine how many contact points are formed with this traverse from start. The blue traverse in Fig. \ref{fig::risk_representation_example2} forms one contact point at the lower left corner of the red obstacle in the middle. Therefore the risk due to number of contact points is 0.03 at state $s_{75}$. It is also worth to note that the orange blocks and purple block also have the color blue. If taking another traverse from the right hand side of the red obstacle to come to the same state $s_{75}$, two contact points (upper right and lower right corner of the middle obstacle) will be formed, instead of one, causing more risk at the same state $s_{75}$. Therefore the entire traverse needs to be considered to determine the risk value associated with number of contact points. 

With all risk values from individual risk elements at individual states computed in Tab. \ref{tab::risk_representation_table}, we can compute the probability of being able to finish each state, shown in the right column ($P(F_i)$). Taking $s_{75}$ as an example again, the probability of finishing $s_{75}$ is the product of the probabilities of all risk elements do not cause failure at this state due to the lateral independence assumption (Eqn. \ref{eqn::risk_representation}): $(1-0.04)\times(1-0.04)\times(1-0.03)=0.89$. In order to finish the path, all the states need to be safely finished. Based on chain rule, the probability of finishing the path is the product of all the entries in the right column: $0.99\times1\times1\times0.96\times0.96\times0.98\times0.89\times95\times0.97\times0.97\times0.94\times0.93=0.62$. Note that in the first row of Tab. \ref{tab::risk_representation_table} probability conditions are omitted in the notation for brevity. Taking the complement will yield the probability of \emph{not} being able to finish the path as $1-0.62=0.38$. This is the risk of the path in Fig. \ref{fig::risk_representation_example2}, meaning if the robot executes this path, there is 0.38 probability that the robot is not able to finish the path.

\begin{figure}[]
\centering
\subfloat[Path 1]{\includegraphics[width=0.5\columnwidth]{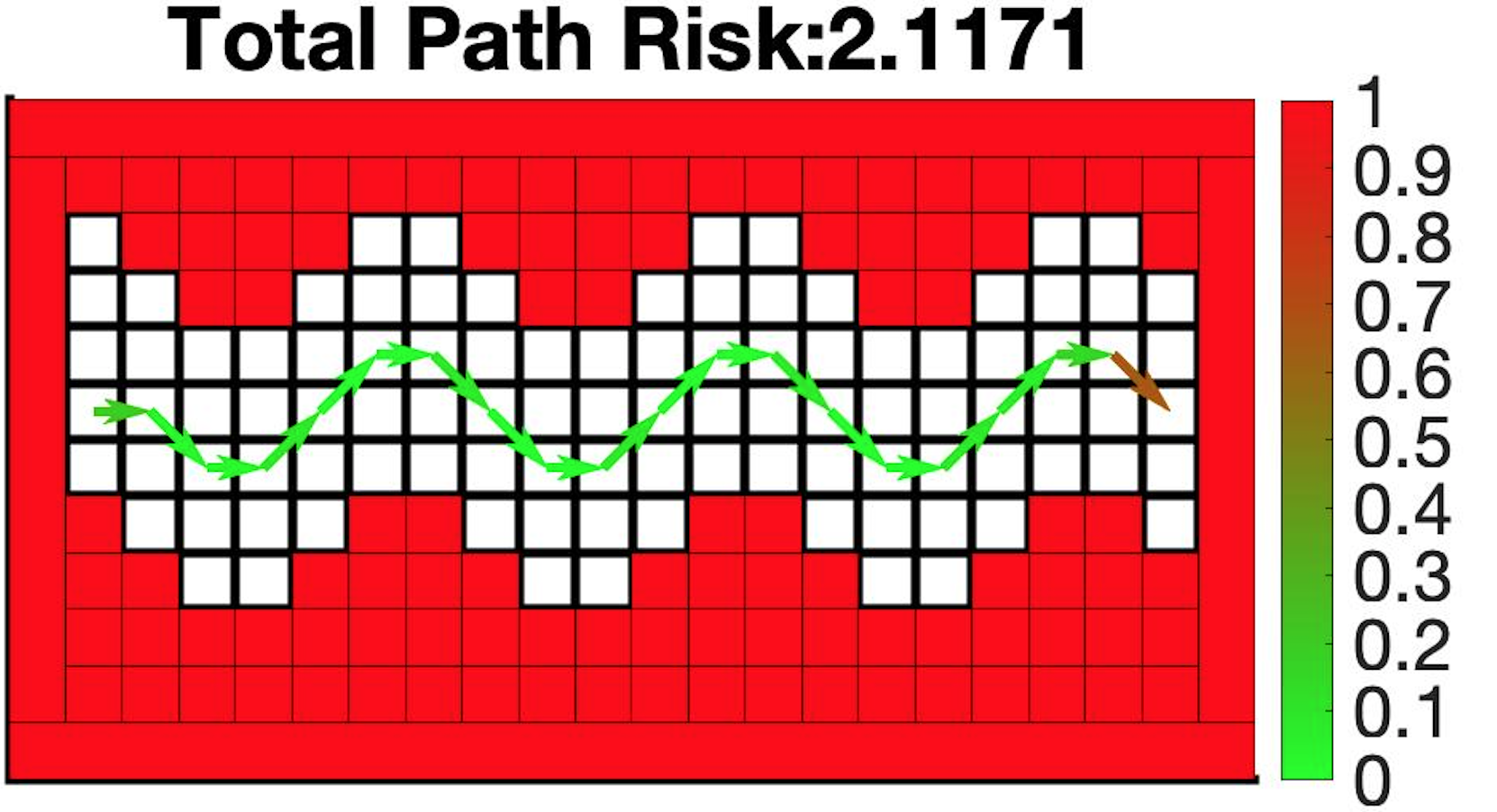}%
\label{fig::path11}}
\hfill
\subfloat[Path 2]{\includegraphics[width=0.5\columnwidth]{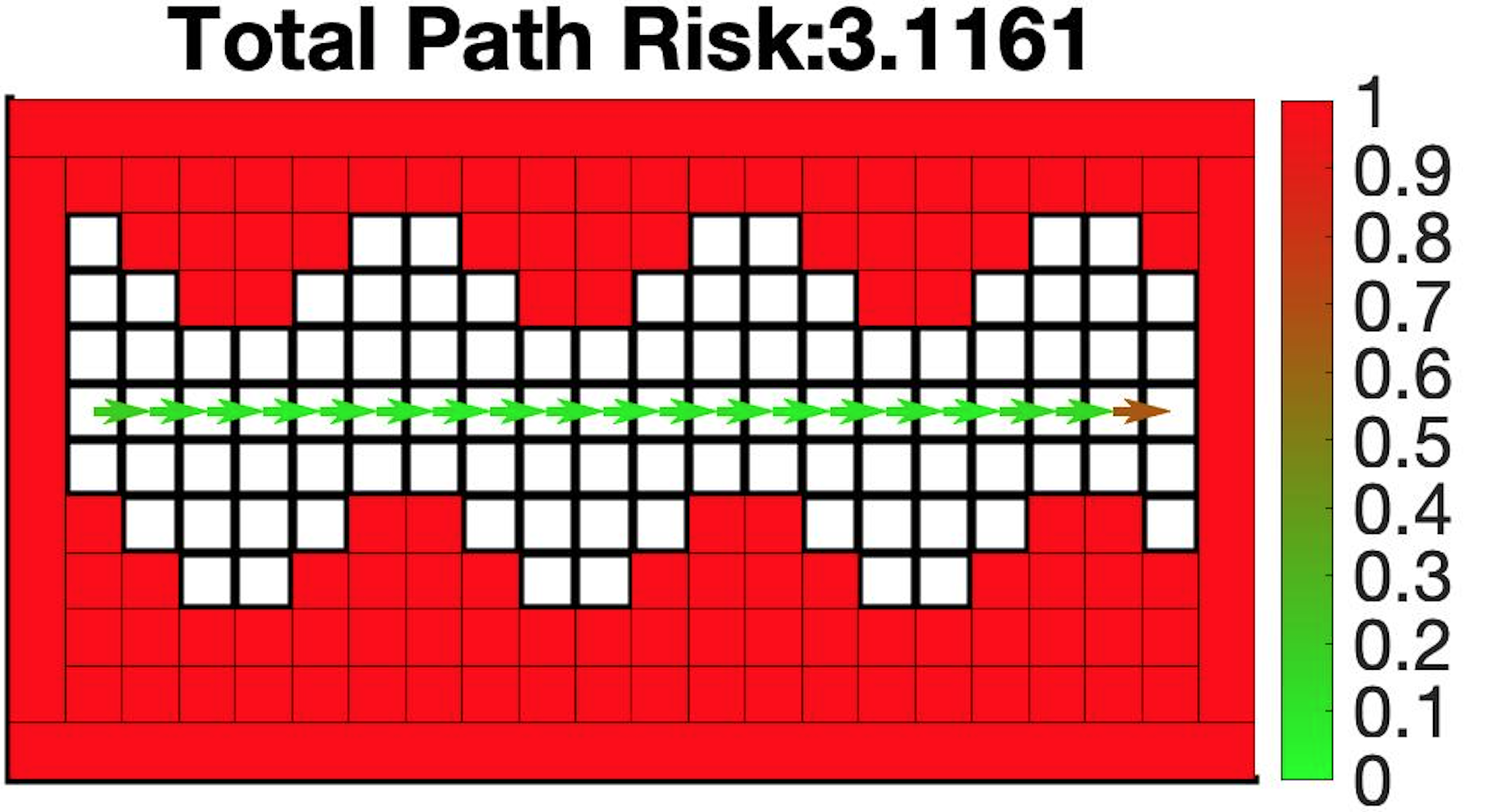}%
\label{fig::path12}}
\caption{Conventional Risk Representation}
\label{fig::conventional}
\end{figure}

\begin{figure}[]
\centering
\subfloat[Path 1]{\includegraphics[width=0.5\columnwidth]{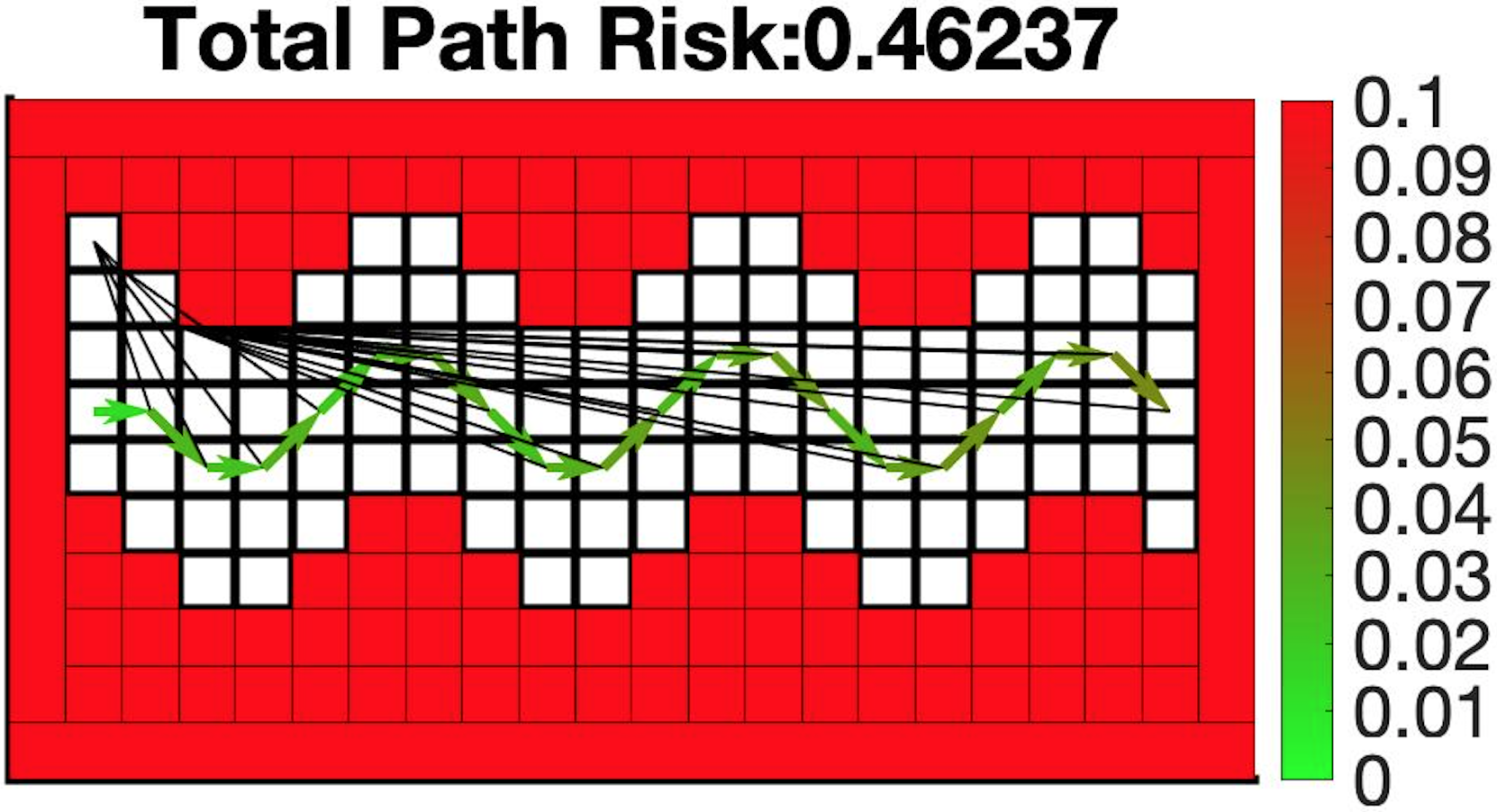}%
\label{fig::path21}}
\hfill
\subfloat[Path 2]{\includegraphics[width=0.5\columnwidth]{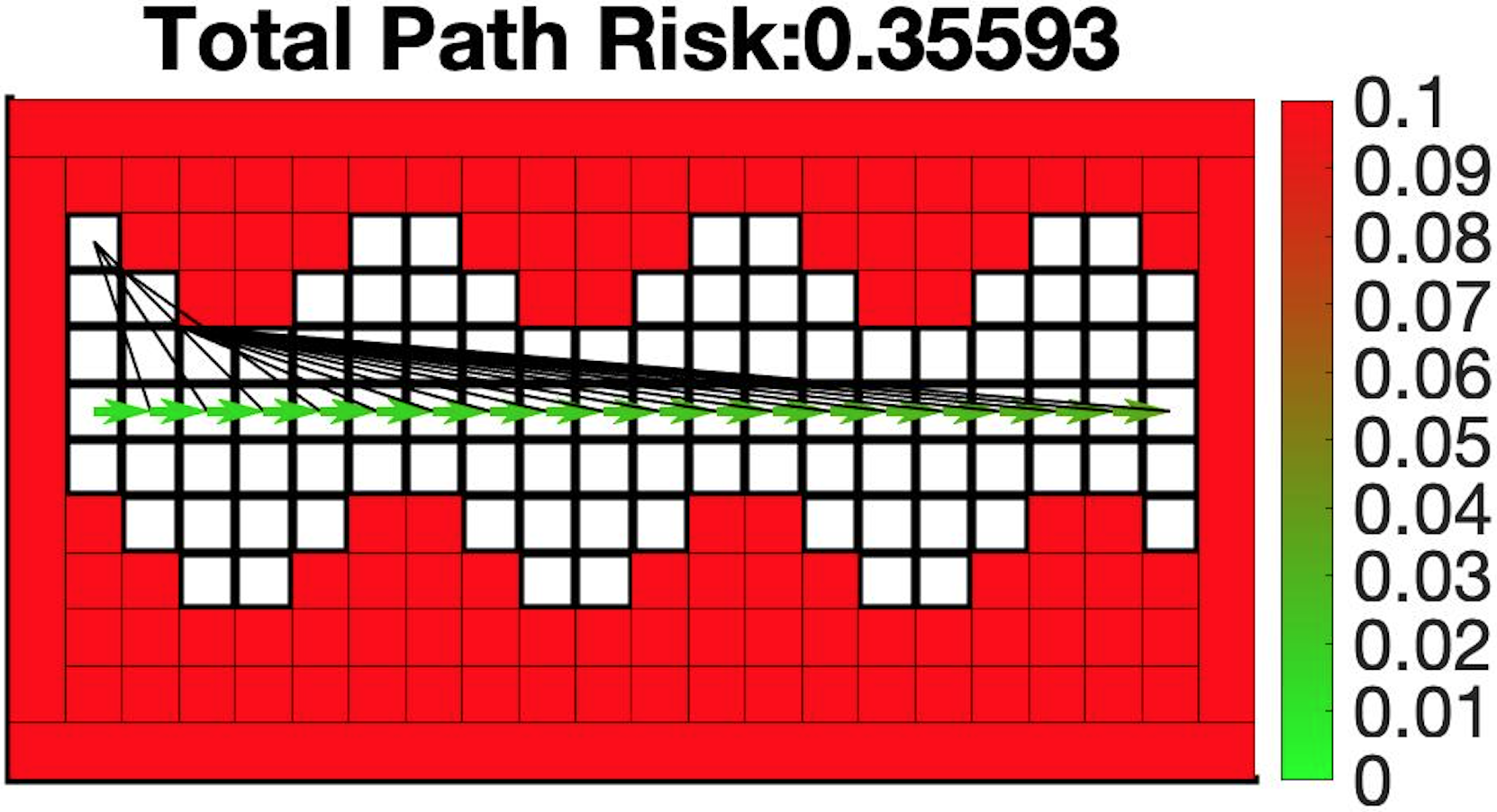}%
\label{fig::path22}}
\caption{Proposed Risk Representation}
\label{fig::proposed}
\end{figure}

Other risk representation examples are shown in Fig. \ref{fig::proposed}. As comparison, results of conventional risk-aware planner based on additive state-dependent risk are presented in Fig. \ref{fig::conventional}. The color of the arrows indicates the risk the robot faces going to each state and the color map is displayed on the right. The robot starts from the left of the map and the goal is going to the right. Six risk elements are chosen as examples from the three risk categories: distance to closest obstacle and visibility from locale-dependent risk elements, action length and turn from action-dependent risk elements, and tether length and number of tether contacts from traverse-dependent risk elements. 

For distance to closest obstacle and visibility, fuzzy logic similar to the approach used in \cite{soltani2004fuzzy} is used. The rationale behind this is the closer a state is to obstacle or the lower visibility the state has, the higher probability that the robot is not able to finish this state. More sophisticated probabilistic model could be used to capture more complex risk relationship. The one important property is locale-dependency. For action length, the risk value is proportional to the norm of the difference between the last and current state. Turning is the difference between two actions, which is further the difference between two states. So second to last, last and current state are taken into account. The risk value is proportional to the norm of the difference between two actions. Again, we assume an easy linear relationship between risk and action length or turning magnitude. Tether length and number of contacts are specific risk elements for tethered robots. Both of them are traverse-dependent and tether planning techniques developed by \citeauthor{xiao2018motion} can output the length and number of contacts. We simply assume risk is proportional to the length and number. 

Fig. \ref{fig::path11} shows the result of conventional additive state-dependent risk representation. Due to the assumption of state-dependency, action length, turn, and tether length, number of contacts cannot be properly addressed by the representation. The only possible risk elements are distance to closest obstacle and visibility, which are evaluated based on state alone. Their risk values at each state are combined using normalization and weighted sum (identical weights for both risk elements) and summed up along the entire path. Fig. \ref{fig::path22} shows the result of the proposed probabilistic risk representation. All six risk elements from all three risk categories could be properly addressed. The risk at each state is now formulated as the probability of the robot not being able to finish the state, displayed in color. The probability of not being able to finish the path, as risk of the path, is computed using propositional logic and probability theory (Eqn. \ref{eqn::risk_representation}).

As shown in Fig. \ref{fig::conventional} and \ref{fig::proposed}, the preference (lower risk) of the twisty path over straight path is switched when using the proposed risk representation instead of the conventional approach. This is because the proposed framework is able to consider more adverse effects due to action-dependent and traverse-dependent risk elements. It is worth to note that the fidelity of the model also depends on the accuracy of the failure probability value of each risk element at each state. How to derive those values more precisely, either using theoretical, empirical, or experimental methods, will be the focus of future work. 

\begin{figure}[]
\centering
\subfloat[Squeezing through]{\includegraphics[width=0.49\columnwidth]{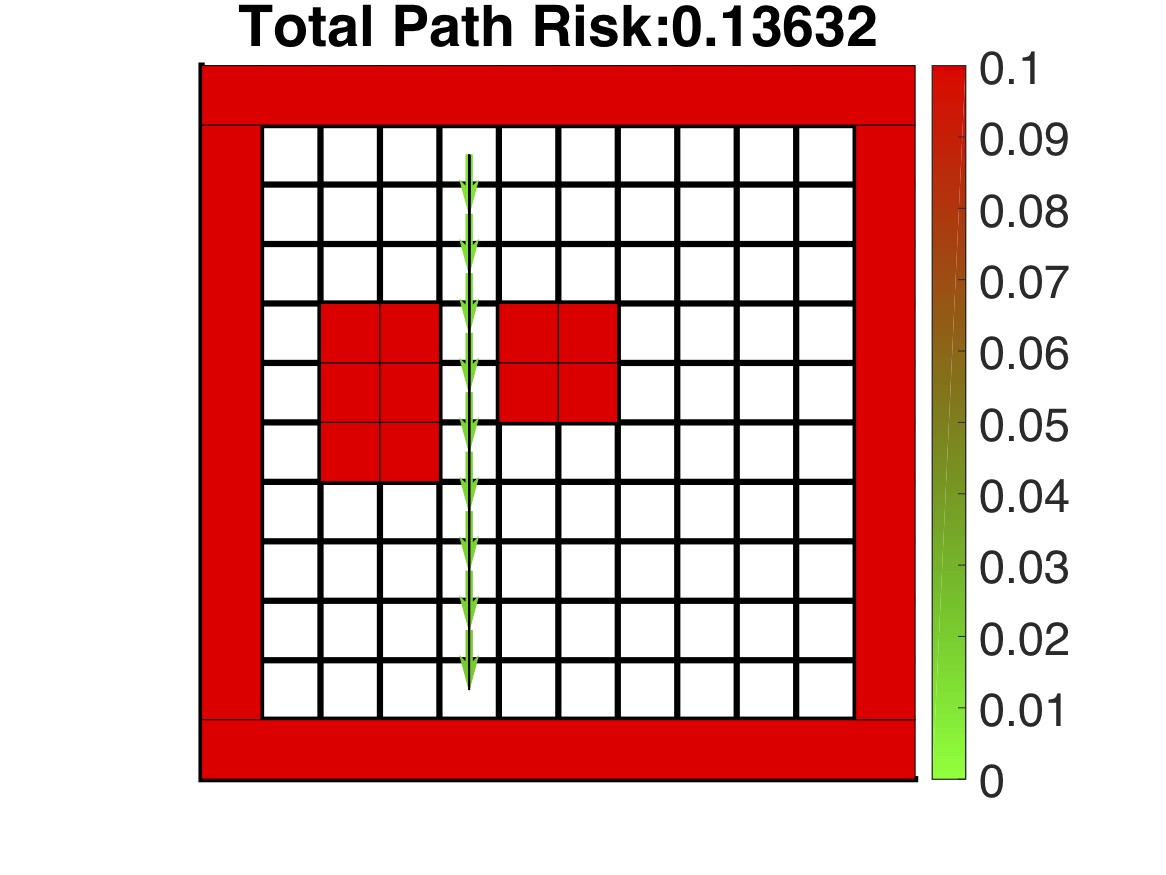}%
\label{fig::in_between}}
\hspace{0.5pt}
\subfloat[Detour]{\includegraphics[width=0.49\columnwidth]{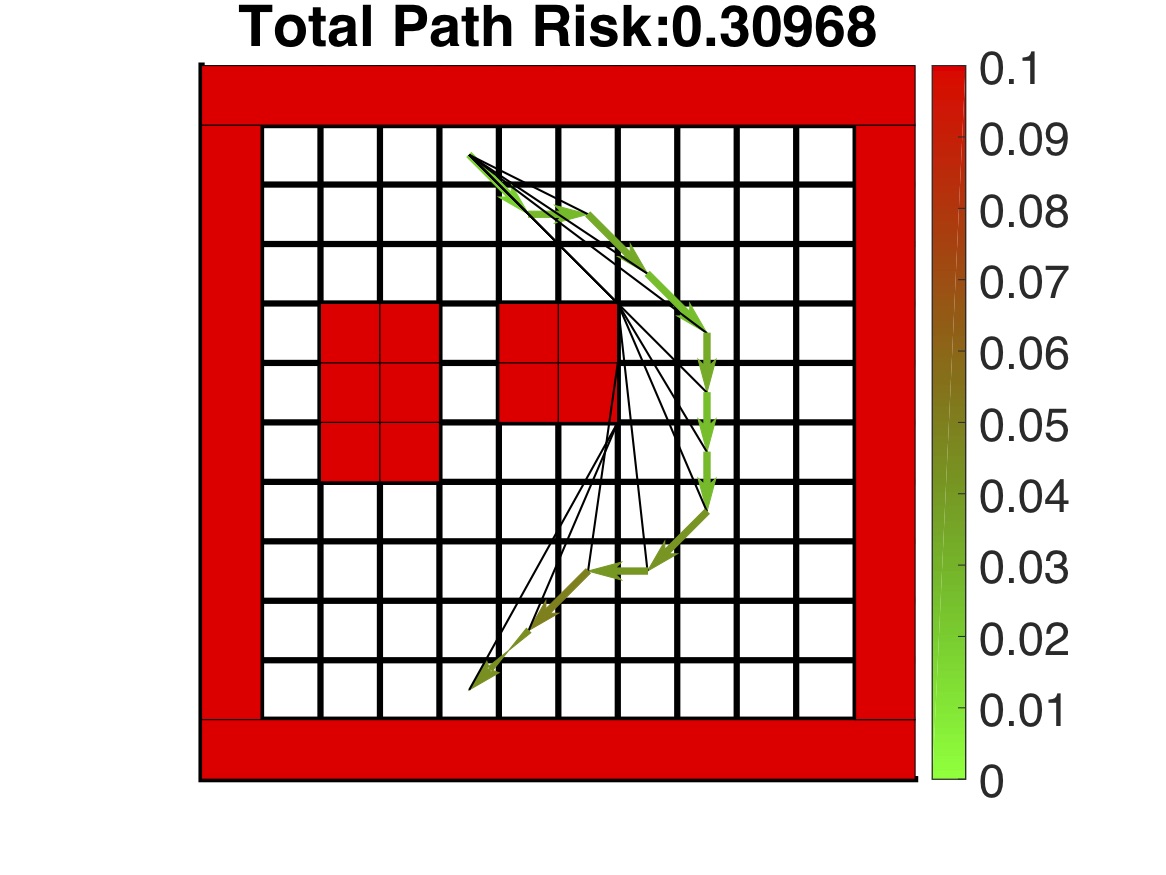}%
\label{fig::around}}
\caption{Path Risk Represented by Proposed Framework}
\label{fig::squeeze_around}
\end{figure}

Another set of examples are shown in Fig. \ref{fig::squeeze_around}. Squeezing through the narrow passage between obstacles compromises distance to closest obstacle and visibility (locale-dependent risk elements), but optimizes action length, turn, tether length, and number of tether contacts. The path, which would be preferred by the conventional approach by only considering state-dependent risk elements (Fig. \ref{fig::around}), actually has a higher probabilistic risk value ($0.31>0.14$). 

\section{Conclusions}
This paper proposes a formal definition of robot motion risk as the probability of the robot not being able to finish the path, which does not exist in the literature. This formal and general definition is applicable to any robotic agents locomoting in unstructured or confined environments. Therefore risk-aware motion is grounded as the motion with maximum probability of being safely finished. An explicit risk representation approach using propositional logic and probability theory is introduced. The use of these formal methods reveals that the risk the robot faces at each state along the path is not only a function of that current state itself, but also dependent on the traverse the robot took from the beginning. By considering not only locale-dependent, but also action-dependent and traverse-dependent risk elements, the proposed risk representation encompasses a comprehensive risk universe in unstructured or confined environments, e.g., motor overheat due to aggressive turning, sensor deterioration due to accumulated environmental interactions, etc., in addition to only obstacle-related risk. It also formally handles their dependencies on the history (longitudinal dependence). The inference using probability chain rule in the proposed risk framework also avoids the ill-founded additive assumption of risk along the entire path and other conservative relaxation techniques in the literature. With a simple lateral independence assumption and the formally reasoned longitudinal dependence, risk is computed as a single probability value of the robot not being able to finish the path. The proposed motion risk framework gives an explicit and intuitive comparison between different motion plans, or paths, for both human and robotic agents. It can be used as a metric to quantify safety for robust robot motion. Future work will focus on developing theoretical or experimental, other than empirical, methods for more precise individual risk values.

\section{ACKNOWLEDGMENT}
This work is supported by NSF 1637955, NRI: A Collaborative Visual Assistant for Robot Operations in Unstructured or Confined Environments. 

\clearpage
\bibliographystyle{aaai}
\bibliography{Reference}

\end{document}